# Mobile Robot Path Planning in Static Environments using Particle Swarm Optimization

M. Shahab Alam, M. Usman Rafique, and M. Umer Khan

*Abstract*— Motion planning is a key element of robotics since it empowers a robot to navigate autonomously. Particle Swarm Optimization is a simple, yet a very powerful optimization technique which has been effectively used in many complex multi-dimensional optimization problems. This paper proposes a path planning algorithm based on particle swarm optimization for computing a shortest collision-free path for a mobile robot in environments populated with static convex obstacles. The proposed algorithm finds the optimal path by performing random sampling on grid lines generated between the robot start and goal positions. Functionality of the proposed algorithm is illustrated via simulation results for different scenarios.

## I. INTRODUCTION

ROBOT motion planning is a vitally important constituent of robotics as it gives a mobile robot the ability and strength to plan its motion and thereby move autonomously between two points for accomplishing different tasks.

Given a robot start and goal position in a 2D environment cluttered with obstacles, the fundamental task of path planning is to generate a shortest path for a robot towards the goal point while avoiding any contact with the obstacles. Based on the volume of information available path planning can be classified into off-line and on-line path planning. In off-line path planning a robot possesses prior information about the environment and in on-line path planning a robot has no prior information about the environment. The general motion planning problem is considered to be NP-hard (non-deterministic polynomial time) because the computational time required for solving such problems increases at an exponential rate with the increase in the size or dimension of the problem [1, 2].

A huge amount of research has been done on motion planning since the pioneering work presented by N. J. Nilsson in late 1960's [3, 4]. Thus far, various motion planning algorithms have been presented by researchers such as Visibility Graphs [5], Voronoi Diagrams [5], Probabilistic Roadmaps [6], Rapidly Exploring Random Trees [7], Potential Fields [8] and many others, each having their strengths and weaknesses.

In order to overcome the problems encountered by the classical motion planning techniques such as high computational cost and time, the use of heuristic techniques for solving the path planning problem has drawn the attention of researchers because of the advantages they offer, such as easy implementation, and fast generation of acceptable solution if there exists one. Particle swarm optimization is a very simple, yet a very powerful heuristic optimization technique which has proved to be very effective in many complex optimization problems [9]. Particle swarm optimization offers many advantages as compared to Genetic Algorithm and other heuristic techniques, such as simplicity, fast convergence, and few parameters that are required to be tuned [10].

This paper proposes a simple path planner based on Particle Swarm Optimization, having the ability to efficiently find an optimal path for a robot in environments strewed with convex obstacles. Efficiency of the proposed algorithm is illustrated via simulation results.

Rest of the paper is structured as follows: Section II discusses Particle Swarm Optimization, Section III presents the problem description, Section IV presents the proposed planner, Section V demonstrates the effectiveness of the proposed planner via simulation results, and finally Section VI gives the conclusion and future recommendations.

## II. PARTICLE SWARM OPTIMIZATION

Particle Swarm Optimization is a Swarm Intelligence based stochastic optimization technique, which was developed in 1995 by James Kennedy and Russell Eberhart [11]. PSO imitates the social behavior of flocks of birds and schools of fish. When a bird or an ant looking for food finds a good path to the food, it instantly transmits the information to the whole swarm and hence rest of the swarm slowly and gradually gravitates towards the food.

In PSO, a swarm of particles is initialized by giving a random position and velocity to each particle in the swarm. These particles are placed in the search space of some problem or function. The fitness function is evaluated with these particles and personal best of each particle is stored in Pbest and global best of the whole swarm is stored in Gbest. In the next iteration these particles are then moved to new positions

M. Shahab Alam is with Department of Mechatronics Engineering, Air University, Main Campus PAF Complex, E-9 Islamabad, Pakistan (e-mail: shahaba@hotmail.com).

M. Usman Rafique is with Department of Mechatronics Engineering, Air University, Main Campus PAF Complex, E-9 Islamabad, Pakistan. (e-mail: m.usman694@gmail.com).

M. Umer Khan is with Department of Mechatronics Engineering, Air University, Main Campus PAF Complex, E-9 Islamabad, Pakistan (e-mail: umer.khan@mail.au.edu.pk).

using (1) and (2). The particles gradually reach the global best positions by communicating the personal best and global best positions to each other. The process repeats until all the particles converge to the same point or until maximum number of iterations is achieved.

PSO uses the following two equations for updating the velocity and position of each particle:

$$V_i = \omega V_{i-1} + c_1 r_1 (P_{best} - X_{i-1}) + c_2 r_2 (G_{best} - X_{i-1}) \quad (1)$$
$$X_i = X_{i-1} + V_i \quad (2)$$

Where $c_1$ and $c_2$ represent the individual and group learning rates respectively, and their values are usually presumed to be equal to 2 [12]. $r_1$ and $r_2$ are uniformly distributed random numbers in the range [0-1]. The parameters $c_1$ and $c_2$ denote the relative importance of the particle's own best position to its neighbor's best position. $\omega$ is the inertia weight factor which is used for improving the search stability. In order to allow the particles to converge more accurately and efficiently the particles' velocities are reduced by using $\omega$. For promoting global exploration of the swarm a larger value of $\omega$ is used and for promoting local exploration a smaller value of $\omega$ is preferred. A balance between local and global exploration can be achieved by a commonly used linearly decreasing inertia weight strategy [13].

$$\omega_{it} = \omega_{max} - \left(\frac{\omega_{max} - \omega_{min}}{it_{max}}\right) it \quad (3)$$

Where $\omega_{max}$ and $\omega_{min}$ are the initial and final values of the inertia weight respectively, and $it_{max}$ is the maximum number of iterations. The values of $\omega_{max} = 0.9$ and $\omega_{min} = 0.4$ are commonly used [13]. Position and velocity of each particle in the swarm is randomly initialized with uniform numbers from $[X_{min}, X_{max}]$ and $[V_{min}, V_{max}]$ respectively.

$$x_i = X_{min} + \sigma_1 (X_{max} - X_{min}) \quad (4)$$
$$v_i = V_{min} + \sigma_2 (V_{max} - V_{min}) \quad (5)$$

Where $\sigma_1$ and $\sigma_2$ represent random numbers from 0 to 1.
PSO can be implemented using the following procedure:

**Algorithm 1: PSO Process**
**Input:** Swarm Size $N$, Maximum Iterations $it\_max$
**Output:** gbest
  **for** each particle $i$ (i=1 to N) **do**
    Initialize position $x_i$
    Initialize velocity $v_i$
  **end for**
  $it \leftarrow 0$
  **while** ($it < it\_max$) **do**
    **for** each particle $i$ (i=1 to N) **do**
      Evaluate fitness $f(x_i)$
      Update position $x_i$ using (1)
      Update velocity $v_i$ using (2)
      Update $pbest_i$ & $gbest$
    **end for**
    $it \leftarrow it + 1$
  **end while**

## III. PROBLEM DESCRIPTION

Robots are required to maneuver in our surroundings, in order to accomplish different tasks. This need makes robot motion planning a very important constituent of robotics.

Assume a rigid robot $\mathcal{A}$ that moves in a 2D Euclidean workspace $\mathcal{W}$ ($\mathcal{W} = \mathbb{R}^2$). Let the workspace $\mathcal{W}$ be populated with rigid obstacles $\mathcal{O}$ where $\mathcal{O}$ denotes the set of all points in $\mathcal{W}$ that lie inside the obstacles i.e. $\mathcal{O} \subseteq \mathcal{W}$. $\mathcal{A}$ is allowed to move in the workspace $\mathcal{W}$, whereas $\mathcal{O}$ remains fixed. Given a start and goal position of $\mathcal{A}$ in $\mathcal{W}$, the objective of path planning is to generate a shortest collision-free path between the start and goal position for $\mathcal{A}$ while avoiding any contact with $\mathcal{O}$.

### A. Environment Modeling

The robot and obstacles from the real world are explicitly transformed into the configuration space (C-Space). A circular mobile robot with radius $r$ is considered such that the robot can translate in the search space without any rotation. The robot is allowed to move in a two-dimensional Euclidean space with static convex obstacles. The shaded polygonal shapes represent obstacles and the white area represents free space for robot movement.

Let $SP$ and $GP$ be the start and goal point of the robot respectively in the co-ordinate system $XOY$ as shown in Fig. 1. In order to reach the goal point $GP$, the robot will need to go through a series of waypoints $wp_n$. First equally divide the straight line joining the start point $SP$ and goal point $GP$ by $n + 1$, where $n$ represents the desired number of waypoints for the robot.

$$Np = d(SP, GP)/n + 1 \quad (6)$$

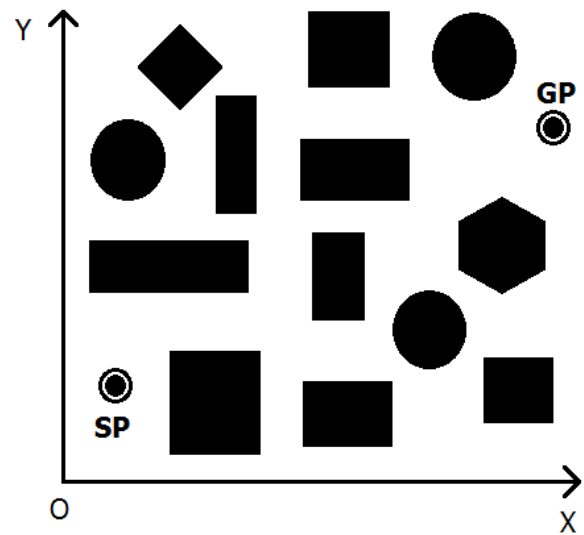

Fig 1. Mobile Robot Workspace

Then draw $n$ number of perpendicular lines on $X$-axis and $Y$-axis each placed at a distance $Np$ as shown in Fig. 2.

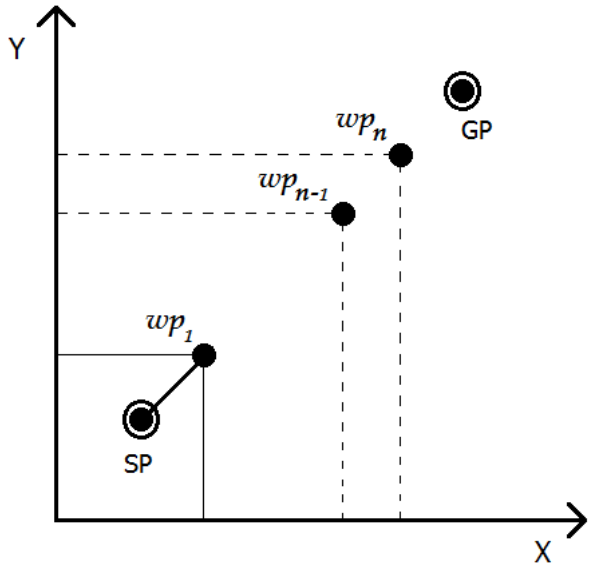

Fig 2. Workspace Model

### A. Objective Function

The objective of path planning is to generate a path or a set of waypoints for a robot from an initial position to a goal position in an environment populated with obstacles while satisfying certain optimization criterion such as shortest distance, minimum time, minimum energy consumption and maximum safety etc.

The performance criterion considered in this research paper is minimizing the path length between the robot start and goal point. The distance between two points $a = (a_1, a_2, ..., a_n)$ and $b = (b_1, b_2, ..., b_n)$ in Euclidean n-space is given by Pythagorean Theorem:

$$d(a,b) = \sqrt{(a_1 - b_1)^2 + (a_2 - b_2)^2 + \cdots + (a_n - b_n)^2} \quad (7)$$

$$d(a,b) = |a - b| = \sqrt{\sum_{i=1}^{n}(a_i - b_i)^2} \quad (8)$$

Hence for seeking a shortest path from $SP(x_{SP}, y_{SP})$ to $GP(x_{GP}, y_{GP})$ via $n$ waypoints $(wp)$ for a robot in a 2-dimensional Euclidean plane can be achieved by the following objective function $\delta$:

$$\delta(SP, GP) = \sqrt{\sum_{i=1}^{n}((x_i - x_{GP})^2 + (y_i - y_{GP})^2)} \quad (9)$$

where

$(x_{SP}, y_{SP})$ : Coordinates of robot start position
$(x_{GP}, y_{GP})$ : Coordinates of robot goal position
$(x_i, y_i)$ : Coordinates of the robot at $i_{th}$ waypoint

### B. Path Planner

The general problem of path planning is to compute a shortest collision-free path for a robot between its initial and final position. After declaring the robot start and goal position, and coordinates of the obstacles, the proposed planner computes the distance between the start and goal point and divides it using (6). The algorithm then generates $n$ perpendicular lines on $X$-axis and $Y$-axis, with each line placed at a distance $Np$. Uniformly distributed random points are generated on these lines using (4).

Then using Algorithm 1, the planner performs random sampling on these grid lines generated between the start and goal positions. Each point is compared with the goal point and the point having minimum distance with the goal point is selected as the feasible waypoint. Points lying inside obstacles are considered as invalid points. The optimal selected waypoints take the robot from $SP$ to $GP$ while satisfying the desired performance criterion. The set of feasible waypoints that links the robot start and goal points represents the robot path $RP$ is given as:

$$RP = \{SP, wp_1, wp_2, ..., wp_n, GP\} \quad (10)$$

A problem arises when the distance between a waypoint $wp_i$ and goal point $GP$ turns out to be equal on either side of an obstacle, hence resulting in discontinuity of the path as shown in Fig. 3. In order to deal with this problem another term β is added to the objective function δ which keeps the waypoints not only close to the goal point but also to each other.

$$\beta = \sqrt{\sum_{i=1}^{n}(wp_{i-1} - wp_i)} \quad (11)$$

Hence the objective function to be minimized can be reformulated as:

$$\delta = \beta + \sqrt{\sum_{i=1}^{n}((x_i - x_{GP})^2 + (y_i - y_{GP})^2)} \quad (12)$$

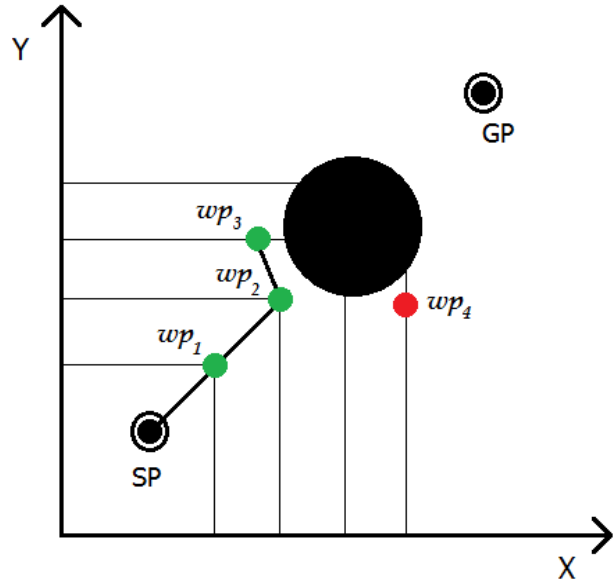

Fig 3. Waypoint Selection

The steps required for implementing the proposed planner are illustrated as follows:

| **Algorithm 2:** Path Planner |
|---|
| **Input:** $SP, GP, n,$ , Coordinates of obstacles  **Output:** Path (Waypoints) from $SP$ to $GP$ |
| Divide the line $\overline{SP - GP}$ into $n + 1$ segments  Generate $n$ perpendicular lines on $X$-axis and $Y$-axis  **repeat**    **for** $i = 1\ to\ n$ **do**       Compute waypoint $wp_i$ using Algorithm. 1    **end for**  **until** $GP$ is reached |

## IV. SIMULATION RESULTS

Simulations of the proposed algorithm are carried out on 1.9 GHz (4 CPUs) Intel® CPU, 2048MB RAM using MATLAB 7.0 R2009b. The following parameter settings are selected: Swarm Size $N = 500$, Maximum Iterations $it_{max} = 100$, Maximum Inertia Weight $\omega_{max} = 0.9$ and Minimum Inertia Weight $\omega_{min} = 0.4$, Maximum Velocity $V_{max} = 200$ and Minimum Velocity $V_{min} = 0$, Social Learning Factor $c_1 = 2$ and Cognitive Learning Factor $c_2 = 2$, Number of Waypoints $n = 100$. Efficiency of the proposed algorithm is tested in several environments having different number of convex obstacles placed at different positions. As robots have some dimensions in real-world, so the sizes of the obstacles are increased by a fixed value in order to keep a safe distance between the robot and the obstacles.

### A. Environment 1

Nine convex obstacles are placed at different positions in environment 1. The robot start point and goal positions are taken to be $SP(0,0)$ and $GP(3.5,9)$ respectively. The red line shown in Fig. 4 represents the optimum path generated by the algorithm and the black shaded polygons represent obstacles.

### B. Environment 2

Environment 2 is populated with seven convex obstacles placed at different positions. The robot start point and goal positions are taken to be $SP(0,0)$ and $GP(7.8,9.2)$ respectively. The proposed algorithm efficiently finds a collision-free path between initial and destination point of the robot. The path generated by the algorithm is represented by red line as shown in Fig. 5.

### C. Environment 3

Environment 3 is strewed with eight static convex obstacles. The robot start point and goal positions are taken to be $SP(0,0)$ and $GP(10,6.5)$ respectively. The red line shown in Fig. 6 represents the optimum path generated by the proposed algorithm.

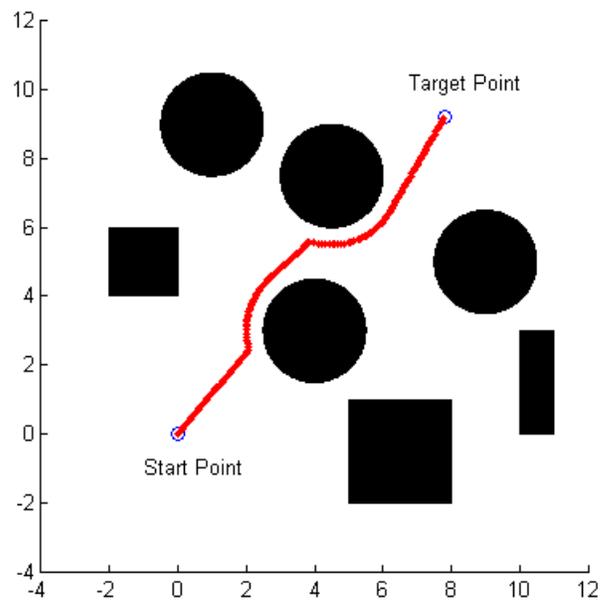

Fig 5. Collision-free Path in Environment 2

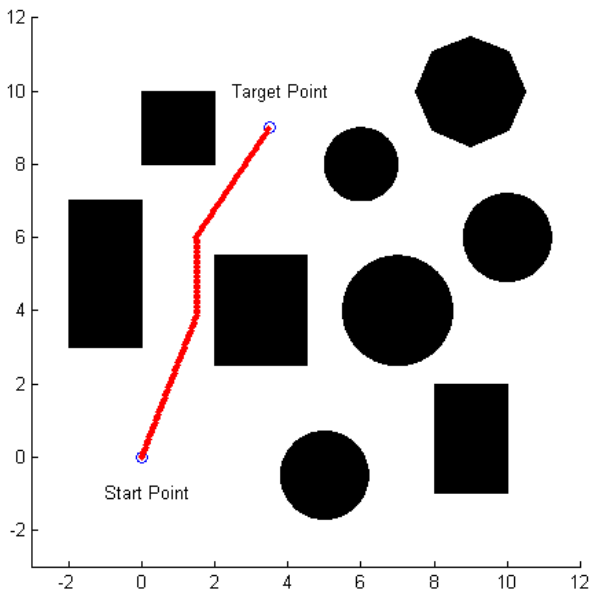

Fig. 4 Collision-free Path in Environment 1

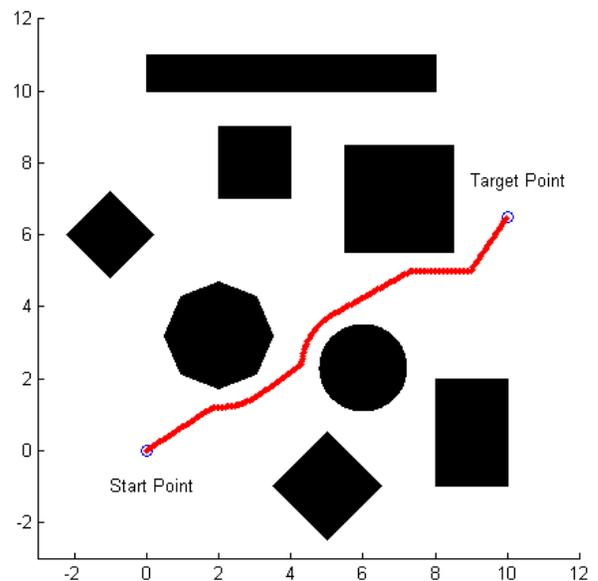

Fig 6 Collision-free Path in Environment 3

*D. Environment 4*

In Environment 4 the robot start point and goal positions are taken to be *SP*(-3,11) and *GP*(8,-2) respectively. The environment is cluttered with sixteen circular shaped obstacles. The optimum path generated by the algorithm is represented by red line as shown in Fig. 7.

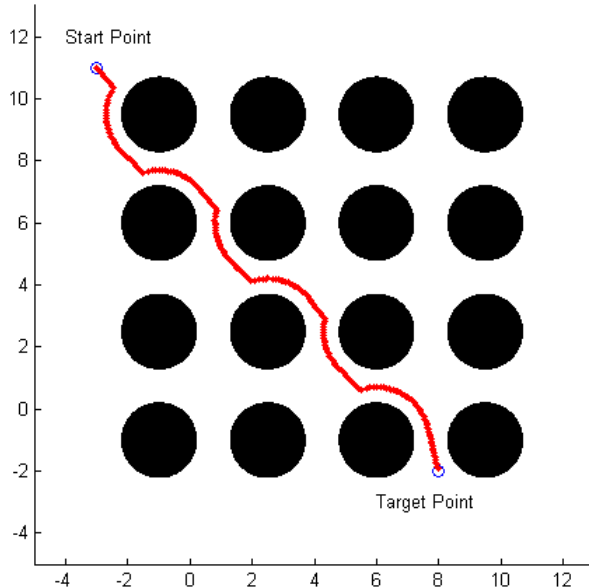

Fig 7. Collision-free Path in Environment 4

## V. CONCLUSION

In this paper robot motion planning problem is treated as an optimization problem and a simple path planner based on particle swarm optimization is presented. The proposed planner performs random sampling on grid lines generated between the robot start and goal positions, and finds the feasible waypoints on these grid lines without high computation and exhaustive search. The optimal waypoints computed by the planner effectively lead the robot to the goal position. Efficiency of the proposed algorithm is demonstrated via simulation results in different environments. This research study will be extended to environments strewed with concave and dynamic obstacles in future.